\documentclass[10pt,twocolumn,letterpaper]{article}

 \usepackage[pagenumbers]{cvpr} % To force page numbers, e.g. for an arXiv version

\usepackage{pifont}% http://ctan.org/pkg/pifont
\newcommand{\cmark}{\ding{51}}%
\newcommand{\xmark}{\ding{55}}%
\usepackage{tabularx}
\usepackage[export]{adjustbox}
\usepackage{multirow}
\usepackage{makecell}
\usepackage{xcolor,colortbl}% http://ctan.org/pkg/xcolor
\usepackage{dsfont}
\usepackage{color} % for cellcolor 
\definecolor{LightGray}{RGB}{230, 230, 230} % A more neutral light gray
\definecolor{DarkGreen}{RGB}{50,150,50} % A darker

\definecolor{MyGreen}{RGB}{0,150,0} % Adjust RGB values as needed
\definecolor{MyRed}{RGB}{200,50,50} % Adjust RGB values as needed

\usepackage{arydshln} % Import the package
\usepackage{subcaption}

\DeclareMathOperator*{\argmin}{arg\,min}
\DeclareMathOperator*{\argmax}{arg\,max}

\definecolor{cvprblue}{rgb}{0.21,0.49,0.74}
\usepackage[pagebackref,breaklinks,colorlinks,allcolors=cvprblue]{hyperref}

\def\paperID{14} % *** Enter the Paper ID here
\def\confName{CVPR}
\def\confYear{2025}

\title{Vocabulary-free few-shot learning for Vision-Language Models}

\author{Maxime Zanella\thanks{\hspace{0.1cm} Equal contributions and corresponding authors. \texttt{\{maxime.zanella,clement.fuchs\}@uclouvain.be}} $^{\hspace{0.5mm} 1,2}$ \hspace{6mm} Clément Fuchs$^{*}$$^{1}$  \hspace{6mm} Ismail Ben Ayed$^{3}$ \hspace{6mm} Christophe De Vleeschouwer$^{1}$
\\
$^{1}$UCLouvain, Belgium \hspace{6mm} $^{2}$UMons, Belgium \hspace{6mm} $^{3}$\'{E}TS Montreal, Canada
}

\begin{document}
\maketitle
\begin{abstract}
Recent advances in few-shot adaptation for Vision-Language Models (VLMs) have greatly expanded their ability to generalize across tasks using only a few labeled examples. However, existing approaches primarily build upon the strong zero-shot priors of these models by leveraging carefully designed, task-specific prompts. This dependence on predefined class names can restrict their applicability, especially in scenarios where exact class names are unavailable or difficult to specify. To address this limitation, we introduce vocabulary-free few-shot learning for VLMs, a setting where target class instances - that is, images - are available but their corresponding names are not. We propose Similarity Mapping (SiM), a simple yet effective baseline that classifies target instances solely based on similarity scores with a set of generic prompts (textual or visual), eliminating the need for carefully handcrafted prompts. Although conceptually straightforward, SiM demonstrates strong performance, operates with high computational efficiency (learning the mapping typically takes less than one second), and provides interpretability by linking target classes to generic prompts. We believe that our approach could serve as an important baseline for future research in vocabulary-free few-shot learning. Code available at \url{https://github.com/MaxZanella/vocabulary-free-FSL}.
\end{abstract}
%learning the mapping with 16384 prompts and 16000 examples takes 0.8s on a Tesla A100
%(learning the mapping with 16384 prompts on 16-shot ImageNet takes 0.8s on a Tesla A100)
\begin{figure}
    \centering
    \includegraphics[width=\linewidth]{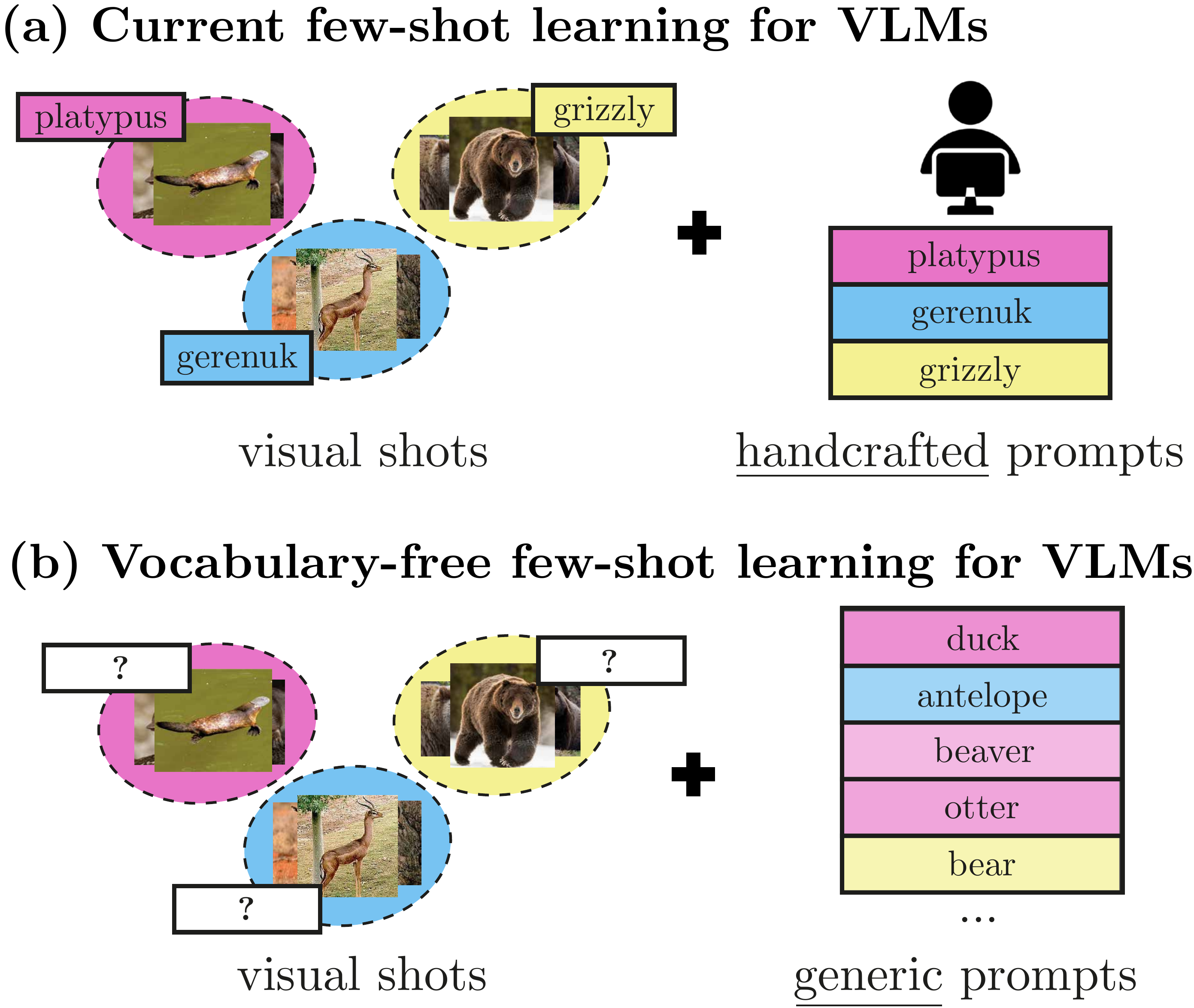}
    \caption{Current few-shot learning methods assume that target class names are known, often requiring manually fine-tuned prompts. In {\em vocabulary-free few-shot learning}, we remove this constraint and rely solely on generic prompts (e.g., derived from ImageNet classes).}
    \label{fig:vocab-free}
\vspace{-1.25pt}
\end{figure}
\section{Introduction}

\par Vision-Language Models (VLMs), such as CLIP~\cite{radford2021learning}, have become essential for cross-modal learning. Such models align images and text through large-scale contrastive training. One of their key strengths is zero-shot classification, which enables them to categorize images solely based on textual prompts that describe target classes. This ability has led to impressive results and motivated recent developments of few-shot learning techniques, which further adapt VLMs to new tasks using only a small number of labeled examples. To achieve this, current few-shot methods build on this strong zero-shot classification, either by incorporating class name tokens into learnable prompts~\cite{zhou2022learning, prograd, khattak2023maple}, by employing handcrafted prompts in combination with adapters~\cite{yu2023task, tip-adapter}, or by using low-rank fine-tuning~\cite{zanella2024low}, to name a few approaches.
\par However, in many cases, class definitions may be vague or ambiguous, making it difficult to design meaningful prompts. Secondly, some classes may require long and complex descriptions that a single prompt may not fully capture, making it preferable to break them down into smaller and more interpretable components. Thirdly, new concepts may emerge, although unknown during VLM pretraining. These practical challenges can hinder direct zero-shot classification and, consequently, limit existing few-shot learning methods that rely on predefined class names. To overcome these limitations, we introduce {\em vocabulary-free few-shot learning}, a framework in which visual instances of the target classes are available, but their corresponding names are not. A comparison with standard few-shot learning for VLMs is provided in Figure~\ref{fig:vocab-free}.

% \par However, knowing the exact classnames is a strong assumption that could be not met in a lot of practical cases (e.g., classes are very vague concepts, classes are new concepts unknown at the time of the VLM pretraining). Another use case is when an exact description of each class would require very long text description, it could be preferable to decompose in smaller prompts/concepts. All this practical cases would hinder current few-shot methods. To this end, we propose a new setting, named {\em vocabulary-free few-shot learning}, in which shots of each target classes are available but not their corresponding names (see Figure \ref{fig:vocab-free}).

\par %In {\em vocabulary-free few-shot learning}, the model has access to groups of images (shots) belonging to the same class but it does not have access to their textual labels.
Without predefined class names, current few-shot methods for VLMs become inapplicable, requiring alternative ways to discriminate classes. In our work, we propose the use of generic textual prompts derived from the ImageNet class names (Table~\ref{tab:main_results}), while illustrating how broader concepts, such as those from the Wordnet lexical database~\cite{miller1995wordnet}, could also be explored. Beyond classification, our approach enhances interpretability by linking target classes to these generic prompts representing meaningful concepts (see Figure~\ref{fig:visu}). This, in turn, may provide a semantic understanding of the target classes. For example, in our experiments, the target class gerenuk is linked to impala and gazelle, other antelope species (Figure \ref{fig:visu_sub1}). 

\par To take advantage of the generic textual prompts, we propose to learn a linear mapping between those prompts and the target classes. Given a small number of labeled examples (shots) per target class, we estimate a mapping that projects similarity scores between images and generic textual prompts onto class assignments. Our approach shares similarities with label mapping techniques used in visual reprogramming~\cite{elsayed2018adversarial, tsai2020transfer, chen2023understanding}, where a mapping function aligns pre-trained model outputs with new task labels. Our method is highly efficient, operates as a black-box model (relying solely on similarity scores rather than direct access to textual or visual embeddings), and requires minimal computational overhead.

\par Beyond its current formulation, we believe that {\em vocabulary-free few-shot learning} and our proposed baseline open up several promising research directions. They include the adaptation of recent advances in few-shot learning to the vocabulary-free problem, the expansion of the set of generic prompts, the integration of richer textual and image-based databases or accounting for prior knowledge about class relationships. Finally, assigning meaningful names to the target classes remains an open problem.

\paragraph{Contributions.} In this work, we introduce {\em vocabulary-free few-shot learning}, a new paradigm where target class visual instances are available, but their exact names are not—challenging the conventional reliance on predefined class names in few-shot Vision-Language Model adaptation. To address this problem, we propose a simple yet effective baseline that learns a linear mapping between generic prompts and target classes, enabling classification without explicit textual target labels. In addition, our approach provides a desirable interpretability property, as the learned mapping offers insights into how target classes relate to known concepts. Finally, we outline several research directions for future improvements, including refining the few-shot learning algorithm. expanding the diversity of generic prompts (texts and/or images), and enabling the fine-grained naming of the target classes.

\begin{figure*}[t]
    \centering
    \includegraphics[width=\linewidth]{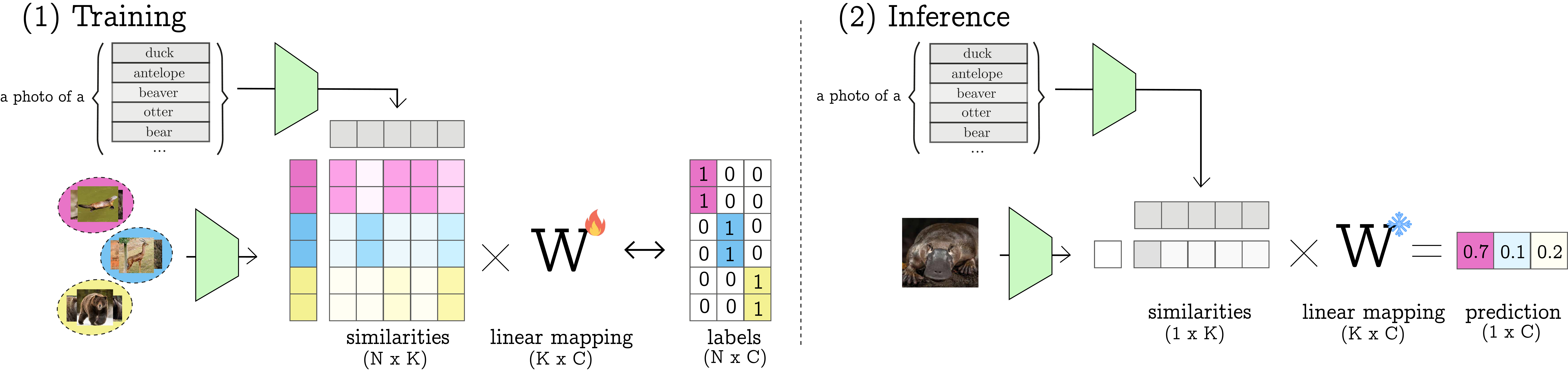}
    \caption{Summary of our approach with three target classes. (1) At training time, a linear operator $W$ is learned via least-squares minimization to map the similarities of $K$ generic prompts and the N visual shots to their corresponding class index (i.e., one-hot encoding). (2) At inference time, the operator $W$ can be used to get a prediction on each test image.}
    \label{fig:framework}
\end{figure*}
\section{Related works}

\paragraph{Few-shot learning in Vision-Language.} Adapting Vision-Language Models (VLMs) to new tasks with minimal supervision has become a predominant area of interest. While their zero-shot capabilities enable broad generalization, fine-tuning on a few labeled examples can still significantly improve performance. Prompt tuning has emerged as a dominant strategy, refining textual and/or visual embeddings to enhance adaptation~\cite{zhou2022learning, khattak2023maple, cocoop, proda, variational, kgcoop, lasp, prograd, plot, khattak2023self}. Methods like CoOp~\cite{zhou2022learning} optimize learnable common continuous tokens attached to the class names, described as a context optimization. MaPLe~\cite{khattak2023maple} extends this strategy by introducing learnable visual tokens in addition to textual ones. A second line of research focuses on adapter-based methods, which modify a small subset of parameters to improve efficiency~\cite{clip-adapter, tip-adapter, yu2023task}. Tip-Adapter~\cite{tip-adapter} leverages memory-based caching, combining stored feature representations with the original zero-shot prediction. TaskRes~\cite{yu2023task} introduces task-specific residual tuning, adapting the initial text embedding of each class prompt. A third path of investigation considers low-rank fine-tuning, exemplified by CLIP-LoRA~\cite{zanella2024low}, which applies low-rank adaptation within both text and visual encoders. All these methods inherently rely on predefined class names, making them unsuitable when explicit class names are unavailable.

\paragraph{Vocabulary-free classification.} Most existing methods for VLMs assume that class names are available at test time, as they are essential for generating textual prompts. However, this assumption becomes impractical when the semantic context is unknown or evolving. Recent works~\cite{NEURIPS2023_619cbddb, conti2024vocabulary} have tackled this issue by assigning names to images from an unconstrained set of semantic concepts. A related approach is retrieval augmented models, based on large-scale image-text pairs database such as LAION-5B~\cite{schuhmann2022laion}, Yfcc100m \cite{thomee2016yfcc100m}, Conceptual Captions \cite{sharma2018conceptual}, or the Public Multimodal Datasets (PMD) \cite{singh2022flava}. These datasets can be used to retrieve similar images and their corresponding captions \cite{long2022retrieval}. Our proposed {\em vocabulary-free few-shot learning} method named SiM differs from these approaches as it focuses on learning a classifier from groups of images (the visual shots) rather than improving single-image classification. 

\paragraph{Label mapping}
Label mapping (LM) has emerged as a core component of visual  reprogramming strategies \cite{elsayed2018adversarial, tsai2020transfer, chen2023understanding}, which focus on reconfiguring a pretrained model for arbitrary downstream tasks using a trainable transformation of the input images, together with an LM function. The latter is needed because the label spaces are often distinct between the pre-training and downstream tasks, which necessitates to map outputs of the pretrained model to downstream labels. Initially, \cite{elsayed2018adversarial} adopted a random mapping strategy, and fully relied on the trainable transformation for accuracy. Later, \cite{tsai2020transfer} introduced frequency label mapping (FLM), which matches the labels of the pre-trained task to each downstream label based on the model’s prediction frequencies. In the context of visual reprogramming, \cite{chen2023understanding} recomputed the LM function as the training of the input transformation progressed, yielding the iterative label mapping algorithm. Recently, \cite{cai2025bayesian} introduced bayesian-guided label mapping (BLM), which provides a bayesian framework for estimating the LM function, leveraging the joint distribution of the pretrained label predictions and ground-truth target classes. As these methods share similarities with our approach in learning a mapping between the pretrained model outputs and target classes, we report their performances in our main experimental results for comparison.

\section{Methodology}
\subsection{Preliminaries}
%\paragraph{Zero-shot classification with VLMs.} 
To fully understand our approach for {\em vocabulary-free few-shot learning} for vision-language models (VLMs), we start by defining the key elements of the usual vision-language classification framework. At its core, CLIP embeds both images and textual descriptions into a common latent space, enabling alignment and measurement of their similarities. Given a set of predefined $C$ classes with known names, the framework relies on generating so-called textual prompts, such as ``a photo of a \{$\text{class}_c$\}". These textual descriptions are first tokenized and then processed by the textual encoder into normalized embeddings ${\mathbf t}_c $ on the unit hypersphere $ \mathcal{S}_{d-1} = \{u \in \mathbb{R}^d ;||u||_2 = 1\}$. Similarly, the image ${\mathbf x}_i$ is transformed by the visual encoder to produce the representation ${\mathbf f}_i \in \mathcal{S}_{d-1}$. This allows for the computation of similarity scores 
\begin{equation} 
l_{i,c} = {\mathbf f}_i^\top {\mathbf t}_c. 
\label{eq:source_similarities}
\end{equation}
Finally, the predicted class for image $x_i$ is obtained by selecting the highest similarity score, i.e. $\hat{c} = \mathrm{argmax}_{c}~l_{i,c}$. Note this approach is not feasible in the vocabulary-free setting since class names \{$\text{class}_c$\} are unavailable, preventing the computation of embeddings $\boldsymbol{t}_c$.

\subsection{Our approach}
Our approach, which we dub Similarities Mapping (SiM),
learns a matrix $W$ to linearly map similarity scores computed with a predefined set of generic prompts to a one-hot representation of the classes, defined by a few-shot set of images.
%involves learning a linear mapping $W$ using the available few-shot set of images and generic prompts, which is then used to classify test images. 
The whole pipeline is summarized in Figure \ref{fig:framework}.
\paragraph{Generic prompts.} Given an image $x_i$, since the embeddings $\boldsymbol{t}_c$ are unavailable, similarities scores $l_{i,c}$ (Eq.~\eqref{eq:source_similarities}) cannot be computed. Rather, the similarity scores are computed with respect to a set of $K$ predefined embeddings $\boldsymbol{t}_k \in \mathcal{S}_{d-1}$, with $1 \leq k \leq K$. $\boldsymbol{t}_k$ can be obtained from a set of arbitrary and generic textual prompts or from other images. The task is then to obtain a class prediction from the scores $l_{i,k}$, given the knowledge of similarity scores for a few-shot set of images $S = \{x_j ; 1 \leq j \leq N\}$. Formally, these similarity scores can be grouped into a matrix
\begin{equation}
L = \begin{pmatrix} 
\mathbf{f}_1^T\mathbf{t}_1 & ... & \mathbf{f}_1^T\mathbf{t}_K \\
... & ... & ... \\
\mathbf{f}_{N}^T\mathbf{t}_1 & ... & \mathbf{f}_{N}^T\mathbf{t}_K
\end{pmatrix} \in \mathbb{R}^{N \times K}. 
\label{eq:sim_matrix}
\end{equation} 
\paragraph{Learning from the few-shot set.} We can then estimate $W~\in~\mathbb{R}^{K \times C}$ such that $Y~\approx~L W$, where $Y \in \mathbb{R}^{N \times C}$ is the matrix whose lines correspond to one-hot encoded labels of the few-shot images. This is achieved by minimizing a least-squares objective with Tikhonov regularization \cite{golub1999tikhonov}, i.e.,
\begin{equation}
W = \argmin_{W} ||Y - L W||_F^2 + \lambda ||W||_F^2,
\label{eq:least_squares}
\end{equation}
$\lambda$ being an hyperparameter. Setting the gradient of the objective in Eq.~\eqref{eq:least_squares} to 0 yields 
\begin{equation}
W = (L^T L + \lambda I_{K})^{-1} Y.
\label{eq:w}
\end{equation}
\paragraph{Classification of test images.} For a test image $x_i$, we get the similarities $l_{i,k} = \mathbf{f}_i^T\mathbf{t}_k$ with respect to the predefined set of arbitrary and generic prompts $\{\mathbf{t}_k\}_{1\leq k\leq K}$. We then use $W$ to map these similarities to the target class scores
\begin{equation}
s_{i,c} = \sum_{k=1}^{K} w_{k,c} l_{i,k} . %= \mathbf{f}_i^T \sum_{k=1}^{K} w_{k,c} \mathbf{t}_{k} .
\label{eq:target_similarities}
\end{equation}
The predicted class is then $\hat{c} = \argmax_c s_{i,c}$. %Note how the similarity score $s_{i,c}$ can be interpreted as a dot product between the input image embeddings and a linear combination of the arbitrary and generic prompts.
By investigating the relative importance of the weights $w_{k,c}$ for a given target class $c$, we can find the embeddings most aligned in the set of $K$ arbitrary and general prompts, potentially providing a high-level semantic understanding of the target class (see Figure~\ref{fig:visu}). %This is illustrated in Figure~\ref{fig:enter-label} as well as in the supplementary material XXXX.

\paragraph{Interpretation of Eq. \eqref{eq:least_squares} as an unsupervised clustering objective.} The problem in Eq. \eqref{eq:least_squares} could be viewed as an unsupervised clustering objective akin to K-means \cite{bauckhage2015k}, but operating on the feature vectors of the visual shots, $(\mathbf{f}_j^T\mathbf{t}_k)_{1 \leq k \leq K}$, $1 \leq j \leq N$, and using a fixed point-to-cluster assignment variables $Y$. The matrix factorization formulation of K-means in \cite{bauckhage2015k} enables to see this perspective. Indeed, the column vectors of matrix $W$, which are the optimization variables, could be viewed as cluster prototypes. So, essentially, our method could be viewed as an unsupervised clustering of the visual shots based on features driven from their similarity with the generic set of prompts.

\section{Experimental validation}
\subsection{Comparative methods}

\paragraph{One-to-One mapping.} This approach seeks to associate a single generic prompt to each target class~$c$, using similarities matrix $L$ (Eq.~\eqref{eq:sim_matrix}). Specifically, to learn the mapping, we adopt the \textit{Frequency Label Mapping} (FLM) algorithm presented in \cite{tsai2020transfer}.%, which greedily constructs the mapping using the number of pairs between the generic prompt with highest similarity and ground truth target class. %Formally, this boils down to computing
% \begin{align}
%     \hat{y}^K_i = \argmax_{y' \in Y^K} (l_{i,k})_{y'},
% \end{align}
%and counting the frequency of ($\hat{y}^K_i = y^K \wedge y^c_i = y^c$).
\paragraph{Bayesian label mapping.} Bayesian label mapping (BLM) was introduced in \cite{cai2025bayesian} and leverages the joint distribution of labels to learn a more flexible many-to-many mapping.
% \begin{align}
%     l_{i,c} = \frac{1}{1}
% \end{align}

\paragraph{Centroids.} To further validate our approach, we compare it to a vision-only few-shot learning baseline inspired by~\cite{snell2017prototypical}. This method follows the usual classification framework of VLMs (see Eq.~\eqref{eq:source_similarities}) but replaces the textual prompts $\boldsymbol{t}_c$ with the sample mean of visual features from images belonging to class $c$. Note that this approach does not make use of the generic prompts and does not provide semantic insights into target classes.
% \begin{align}
%     w_{c} = \frac{\sum_i f_i \cdot \mathbb{1}\{y_i = c\}}{\sum_i \mathbb{1}\{y_i = c\}} ; \quad  \hat{c} = \argmax_c f_i^T w_c
% \end{align}

\paragraph{Standard few-shot learning for VLMs.} Additionally we provide results for few-shot methods requiring the class names. Because of the large popularity of these fields, we report some of the most popular methods i.e., the text prompt tuning method CoOp \cite{zhou2022learning}, the vision and text prompt tuning MaPLe \cite{khattak2023maple}, the cache-based method Tip-Adapter \cite{tip-adapter}, the adapter-based method TaskRes \cite{yu2023task} and the low-rank adaptation method CLIP-LoRA \cite{zanella2024low}. These methods serve as upper-bound references in our experiments, as they benefit from access to predefined class names and can therefore leverage zero-shot prediction, unlike vocabulary-free approaches.

\subsection{Experimental setting}
\paragraph{Datasets.} We adapt the setting of previous works \cite{zhou2022learning} and use 11 datasets for image classification tasks, namely ImageNet~\cite{imagenet} a large-scale benchmark for object recognition, SUN397~\cite{sun397} for fine-grained classification of scenes,  Aircraft \cite{aircraft} for classification of aircraft types, EuroSAT~\cite{eurosat} for satellite imagery, StanfordCars~\cite{cars} for car models, Food101~\cite{food} for food items, Pets \cite{pets} for pet types, Flower102~\cite{flower} for flower species, Caltech101~\cite{caltech101} for a variety of general objects, DTD~\cite{dtd} for texture types and UCF101~\cite{ucf101} for action recognition.
\paragraph{ImageNet class names for prompts.} For the main results, we use the classes of ImageNet \cite{imagenet} to generate $K = 1000$ prompts of the form ``a photo of a \{$\text{class}_k$\}.", and subsequently obtain the associated embeddings $\boldsymbol{t}_k$. 
\paragraph{ImageNet images for prompts.} To demonstrate that similarities can be computed using a wide variety of prompts, we investigate an alternative setting where image-based representations replace textual prompts. As discussed later, this highlights that improving the choice of prompts could be an interesting direction for future research. In this setting, for each random seed we randomly draw a single image for each of the $K=1000$ classes, which are then processed by the visual encoder to yield the normalized embeddings $\boldsymbol{t}_k$. Hence, the similarities in the matrix $L$ are images to images rather than images to text.
\paragraph{Wordnet vocabulary for prompts.} In this setting, we select all words in Wordnet \cite{miller1995wordnet} which are related to at least one of the words in [``building", ``vehicle", ``food", ``flower", ``animal", ``texture", ``action", ``furniture"], resulting in a vocabulary of $K=16452$ words. We then generate prompts of the form ``a photo of a \{$\text{word}_k$\}." which are processed by the text encoder and normalized to yield the embeddings $\boldsymbol{t}_k$.
%Subsequently, for each dataset, we randomly sample $|S|$ images in the relevant few-shot setting from the training split, and compute the similarity scores matrix $L$. The latter is used to learn the linear mapping $W$ according to Equation \ref{eq:w}. Finally, we predict on the complete testing set and compute evaluation metrics. Results are averaged over 3 random seeds.
\begin{table*}[t]
\caption{Detailed results for the 10 datasets for two CLIP backbones. Top-1 accuracy averaged over 3 random seeds is reported. Highest value for Vocabulary-free methods is highlighted in \textbf{bold}.}
\label{tab:main_results}
\centering
\begin{subtable}{0.85\linewidth}
\caption{Results for the CLIP ViT-B/16 backbone.}
\label{tab:main_results_vitb16}
\resizebox{\textwidth}{!}{%
\begin{tabular}{llcccccccccccc}
\toprule
 & Method & Vocabulary-free & SUN & Aircraft & EuroSAT & Cars & Food & Pets &  Flowers & Caltech & DTD & UCF & Average
\\ \midrule 
 & CLIP  & \textcolor{MyRed}{\xmark} & 62.6 & 24.7 & 47.5 & 65.3 & 86.1 & 89.1 & 71.4 & 92.9 & 43.6 & 66.7 &  65.0 \\

\midrule
\midrule
\multirow{8}{*}{\rotatebox{90}{4-shot}}
& CoOp  & \textcolor{MyRed}{\xmark} & 69.7 & 30.9 & 69.7 & 74.4 & 84.5 & 92.5  & 92.2 & 94.5 & 59.5 & 77.6 &  74.6 \\

 & TIP-Adapter-F  & \textcolor{MyRed}{\xmark} & 70.8 & 35.7 & 76.8 & 74.1 & 86.5 & 91.9 & 92.1 & 94.8 & 59.8 & 78.1 & 76.1 \\

& TaskRes & \textcolor{MyRed}{\xmark} & 72.7 & 33.4 & 74.2 & 76.0 & 86.0 & 91.9 & 85.0 & 95.0 & 60.1 & 76.2 &  75.1\\
 & MaPLe & \textcolor{MyRed}{\xmark} & 71.4 & 30.1 & 69.9 & 70.1 & 86.7 & 93.3 & 84.9 & 95.0 & 59.0 & 77.1 & 73.8 \\

 & CLIP-LoRA   & \textcolor{MyRed}{\xmark} & 72.8 & 37.9 & 84.9 & 77.4 & 82.7 & 91.0 & 93.7 & 95.2 & 63.8 & 81.1 &  78.1\\
 \cdashline{2-14} % Dashed line spanning columns 1 to 2
 & One-to-One & \textcolor{MyGreen}{\cmark} & 20.5 & 2.3 & 27.9 & 4.3 & 16.9 & 68.2 & 11.2 & 76.3 & 23.8 & 39.7 & 29.1 \\ 

  & BLM & \textcolor{MyGreen}{\cmark} & 26.4 & 1.6 & 38.1 & 2.7 & 19.0 & 63.4 & 17.5 & 81.9 & 34.1 & 45.1 & 33.0 \\

& centroids & \textcolor{MyGreen}{\cmark} & 60.9 & 32.4 & 70.1 & 57.8 & 71.0 & 64.6 & \textbf{91.2} & 91.3 & 53.9 & 70.9 & 66.4 \\

 &  \cellcolor{LightGray}SiM (ours) & \cellcolor{LightGray}\textcolor{MyGreen}{\cmark} & \cellcolor{LightGray}\textbf{62.7} & \cellcolor{LightGray}\textbf{33.2} & \cellcolor{LightGray}\textbf{75.8} & \cellcolor{LightGray}\textbf{60.5} & \cellcolor{LightGray}\textbf{75.4} & \cellcolor{LightGray}\textbf{79.2} & \cellcolor{LightGray}89.7 & \cellcolor{LightGray}\textbf{93.2} & \cellcolor{LightGray}\textbf{59.0} & \cellcolor{LightGray}\textbf{73.8} & \cellcolor{LightGray}\textbf{70.2} \\

\midrule
\multirow{8}{*}{\rotatebox{90}{8-shot}}
& CoOp  & \textcolor{MyRed}{\xmark} & 71.9 & 38.5 & 77.1 & 79.0 & 82.7 & 91.3 & 94.9 & 94.5 & 64.8 & 80.0 &  77.5 \\

 & TIP-Adapter-F  & \textcolor{MyRed}{\xmark} & 73.5 & 39.5 & 81.3 & 78.3 & 86.9 & 91.8 & 94.3 & 95.2 & 66.7 & 82.0 &  79.0 \\

& TaskRes  & \textcolor{MyRed}{\xmark} & 74.6 & 40.3 & 77.5 & 79.6 & 86.4 & 92.0 & 96.0 & 95.3 & 66.7 & 81.6 & 79.0 \\
 & MaPLe  & \textcolor{MyRed}{\xmark} & 73.2 & 33.8 & 82.8 & 71.3 & 87.2 & 93.1 & 90.5 & 95.1 & 63.0 & 79.5 & 77.0\\

& CLIP-LoRA &  \textcolor{MyRed}{\xmark} & 74.7 & 45.7 & 89.7 & 82.1 & 83.1 & 91.7 & 96.3 & 95.6 & 67.5 & 84.1 &  81.1 \\
\cdashline{2-14} % Dashed line spanning columns 1 to 2
& One-to-One & \textcolor{MyGreen}{\cmark} & 22.6 & 2.4 & 30.2 & 4.5 & 18.5 & 69.4 & 12.8 & 77.2 & 28.5 & 41.3 & 30.7 \\

& BLM & \textcolor{MyGreen}{\cmark} & 28.4 & 1.9 & 41.2 & 2.6 & 21.4 & 59.4 & 19.1 & 86.2 & 39.7 & 45.1 & 34.5 \\

& centroids & \textcolor{MyGreen}{\cmark} & 66.8 & 36.9 & 74.3 & 65.7 & 77.6 & 74.6 & \textbf{94.1} & 92.7 & 60.4 & 76.2 & 71.9 \\

 & \cellcolor{LightGray}SiM (ours) & \cellcolor{LightGray}\textcolor{MyGreen}{\cmark} & \cellcolor{LightGray}\textbf{67.2} & \cellcolor{LightGray}\textbf{38.4} & \cellcolor{LightGray}\textbf{80.1} & \cellcolor{LightGray}\textbf{68.9} & \cellcolor{LightGray}\textbf{80.1} & \cellcolor{LightGray}\textbf{84.6} & \cellcolor{LightGray}92.8 & \cellcolor{LightGray}\textbf{94.8} & \cellcolor{LightGray}\textbf{64.5} & \cellcolor{LightGray}\textbf{77.0} & \cellcolor{LightGray}\textbf{74.9} \\
\midrule
\multirow{8}{*}{\rotatebox{90}{16-shot}}

& CoOp   & \textcolor{MyRed}{\xmark} & 74.9 & 43.2 & 85.0 & 82.9 & 84.2 & 92.0 & 96.8 & 95.8 & 69.7 & 83.1 &  80.8 \\

 & TIP-Adapter-F & \textcolor{MyRed}{\xmark} &  76.0 & 44.6 & 85.9 & 82.3 & 86.8 & 92.6 & 96.2 & 95.7 & 70.8 & 83.9 & 81.5 \\

& TaskRes  & \textcolor{MyRed}{\xmark} & 76.1 & 44.9 & 82.7 & 83.5 & 86.9 & 92.4 & 97.5 & 95.8 & 71.5 & 84.0 &  81.5 \\
 & MaPLe  & \textcolor{MyRed}{\xmark} & 74.5 & 36.8 & 87.5 & 74.3 & 87.4 & 93.2 & 94.2 & 95.4 & 68.4 & 81.4 & 79.3 \\

 & CLIP-LoRA & \textcolor{MyRed}{\xmark} & 76.1 & 54.7 & 92.1  & 86.3 & 84.2 & 92.4 & 98.0 & 96.4 & 72.0 & 86.7 &  83.9 \\
 \cdashline{2-14} % Dashed line spanning columns 1 to 2
 & One-to-One & \textcolor{MyGreen}{\cmark} & 24.0 & 2.4 & 34.2 & 4.9 & 18.9 & 71.9 & 14.0 & 78.0 & 30.9 & 43.4 & 32.3 \\

 & BLM & \textcolor{MyGreen}{\cmark} & 29.3 & 1.1 & 36.0 & 2.5 & 21.3 & 49.6 & 20.1 & 86.1 & 44.3 & 47.8 & 33.8 \\

& centroids & \textcolor{MyGreen}{\cmark} & \textbf{70.1} & 40.7 & 76.6 & 71.6 & 80.9 & 78.8 & \textbf{95.5} & 93.8 & 63.1 & 77.4 & 74.8 \\

 & \cellcolor{LightGray}SiM (ours) & \cellcolor{LightGray}\textcolor{MyGreen}{\cmark} & \cellcolor{LightGray}\textbf{70.1} & \cellcolor{LightGray}\textbf{43.7} & \cellcolor{LightGray}\textbf{85.6} & \cellcolor{LightGray}\textbf{74.8} & \cellcolor{LightGray}\textbf{82.8} & \cellcolor{LightGray}\textbf{88.1} & \cellcolor{LightGray}\textbf{95.5} &\cellcolor{LightGray}\textbf{95.3} & \cellcolor{LightGray}\textbf{69.9} & \cellcolor{LightGray}\textbf{79.0} & \cellcolor{LightGray}\textbf{78.5} \\
\bottomrule
\end{tabular}}

\vspace{0.5cm}

\end{subtable}
\begin{subtable}{0.85\linewidth}
\caption{Results for the CLIP ViT-L/14 backbone.}
\label{tab:main_results_vitl14}
\resizebox{\textwidth}{!}{%
\begin{tabular}{llcccccccccccc}
\toprule
 & Method & Vocabulary-free & SUN & Aircraft & EuroSAT & Cars & Food & Pets &  Flowers & Caltech & DTD & UCF & Average
\\ \midrule 
 & CLIP  & \textcolor{MyRed}{\xmark} &  67.6 & 32.6 & 58.0 & 76.8 & 91.0 & 93.6 & 79.4 & 94.9 & 53.6 & 74.2 & 72.2 \\

\midrule
\midrule
\multirow{8}{*}{\rotatebox{90}{4-shot}}
& CoOp   & \textcolor{MyRed}{\xmark} &  73.7 & 41.8 & 77.9 & 82.6 & 88.8 & 94.7 & 94.9 & 96.1 & 64.3 & 83.6 & 79.8\\

& TIP-Adapter-F  & \textcolor{MyRed}{\xmark} & 74.1 & 47.4 & 81.4 & 82.3 & 91.2 & 94.0 &  95.5 & 96.5 & 64.4 & 83.9 & 81.1 \\

& TaskRes & \textcolor{MyRed}{\xmark} & 74.9 & 42.5 & 76.6 & 83.6 & 90.7 & 94.4 & 90.3 & 96.5 & 65.4 & 80.1 & 79.5\\

& MaPLe & \textcolor{MyRed}{\xmark}  & 76.0 & 40.4 & 74.6 & 80.3 & 91.5 & 95.0 & 93.2 & 97.0 & 64.5 & 82.8 & 79.5 \\

 & CLIP-LoRA   & \textcolor{MyRed}{\xmark} & 76.7 &  48.9 & 86.4 & 85.2 & 89.6 & 93.9 & 97.4 &  97.2 & 70.4 & 86.4 & 83.2\\
 \cdashline{2-14} % Dashed line spanning columns 1 to 2
 & One-to-One & \textcolor{MyGreen}{\cmark} & 23.7 & 2.6 & 37.2 & 5.1 & 17.7 & 71.3 & 15.9 & 73.8 & 22.5 & 40.4 & 31.0 \\

  & BLM & \textcolor{MyGreen}{\cmark} & 29.8 & 2.6 & 45.6 & 4.2 & 21.3 & 68.6 & 21.6 & 86.3 & 33.4 & 49.7 & 36.3 \\

 & centroids & \textcolor{MyGreen}{\cmark} & 66.1 & 42.5 & 76.0 & 69.4 & 82.0 & 77.7 & \textbf{96.3} & 94.4 & 58.4 & 78.7 & 74.2 \\
 & \cellcolor{LightGray}SiM (ours) & \cellcolor{LightGray}\textcolor{MyGreen}{\cmark} & \cellcolor{LightGray}\textbf{67.2} & \cellcolor{LightGray}\textbf{43.0} & \cellcolor{LightGray}\textbf{80.9} & \cellcolor{LightGray}\textbf{72.5} & \cellcolor{LightGray}\textbf{85.6} & \cellcolor{LightGray}\textbf{87.3} & \cellcolor{LightGray}\textbf{96.3} & \cellcolor{LightGray}\textbf{95.8} & \cellcolor{LightGray}\textbf{61.4} & \cellcolor{LightGray}\textbf{81.8} & \cellcolor{LightGray}\textbf{77.2} \\

\midrule
\multirow{8}{*}{\rotatebox{90}{8-shot}}
& CoOp  & \textcolor{MyRed}{\xmark} & 75.5 & 48.7 & 81.4 & 85.9 & 89.2 & 94.5 & 97.5 & 96.5 & 68.8 & 86.0 &  82.4\\

& TIP-Adapter-F  &  \textcolor{MyRed}{\xmark} & 76.7 &  50.4 & 84.9 & 85.9 & 91.4 & 94.1 & 97.3 & 96.9 & 71.2 &  86.2 & 83.5\\
& TaskRes  & \textcolor{MyRed}{\xmark} &  76.0 & 51.1 & 81.1 & 85.7 & 91.1 & 94.5 & 96.7 & 96.9 & 69.4 & 85.6 & 82.8\\
& MaPLe  & \textcolor{MyRed}{\xmark} & 77.2 & 42.9 & 80.7 & 81.8 & 90.1 & 95.0 & 95.8 & 96.8 & 69.5 & 85.1 & 81.5 \\
& CLIP-LoRA &  \textcolor{MyRed}{\xmark} & 78.0 & 57.5 & 90.0 & 88.7 & 89.7 &  94.2 & 98.0 & 97.0 & 72.2 & 88.3 &  85.4 \\
\cdashline{2-14} % Dashed line spanning columns 1 to 2
& One-to-One & \textcolor{MyGreen}{\cmark} & 25.5 & 2.8 & 42.4 & 5.2 & 19.9 & 70.6 & 16.3 & 76.8 & 26.1 & 43.7 & 32.9 \\

& BLM & \textcolor{MyGreen}{\cmark} & 32.1 & 1.5 & 52.8 & 3.2 & 23.3 & 63.4 & 25.4 & 89.2 & 39.3 & 50.4 & 38.0 \\

& centroids & \textcolor{MyGreen}{\cmark} &  \textbf{71.9} & 47.0 & 79.8 & 76.8 & 86.3 & 85.6 & 97.4 & 95.6 & 65.3 & 82.4 & 78.8 \\
 & \cellcolor{LightGray}SiM (ours) & \cellcolor{LightGray}\textcolor{MyGreen}{\cmark} & \cellcolor{LightGray}71.6 & \cellcolor{LightGray}\textbf{47.6} & \cellcolor{LightGray}\textbf{85.2} & \cellcolor{LightGray}\textbf{79.8} & \cellcolor{LightGray}\textbf{88.4} & \cellcolor{LightGray}\textbf{91.6} & \cellcolor{LightGray}\textbf{97.5} & \cellcolor{LightGray}\textbf{96.9} & \cellcolor{LightGray}\textbf{68.3} & \cellcolor{LightGray}\textbf{83.3} & \cellcolor{LightGray}\textbf{81.0} \\

\midrule
\multirow{8}{*}{\rotatebox{90}{16-shot}}

& CoOp  & \textcolor{MyRed}{\xmark} & 77.9 & 53.0 & 86.7 & 87.4 & 90.2 & 94.5 & 98.6 & 97.5 & 73.7 & 86.7 & 84.6\\
& TIP-Adapter-F &  \textcolor{MyRed}{\xmark} & 79.6 & 55.8 & 86.1 & 88.1 & 91.6 & 94.6 & 98.3 & 97.5 & 74.0 & 87.4 & 85.3 \\
& TaskRes  & \textcolor{MyRed}{\xmark} &  76.9 & 55.0 & 84.3 & 87.6 & 91.5 & 94.7 & 97.8 & 97.3 & 74.4 & 86.6 & 84.6 \\
& MaPLe  & \textcolor{MyRed}{\xmark} & 78.8 & 46.3 & 85.4 & 83.6 & 92.0 & 95.4 & 97.4 & 97.2 & 72.7 & 86.5 & 83.5 \\
 & CLIP-LoRA & \textcolor{MyRed}{\xmark} & 79.4 & 66.2 & 93.1 & 90.9 & 89.9 & 94.3 & 99.0 & 97.3 & 76.5 & 89.9 & 87.7 \\
 \cdashline{2-14} % Dashed line spanning columns 1 to 2
 & One-to-One & \textcolor{MyGreen}{\cmark} & 26.7 & 2.9 & 47.1 & 5.9 & 20.5 & 72.8 & 18.6 & 77.1 & 29.4 & 43.6 & 34.4 \\

 & BLM & \textcolor{MyGreen}{\cmark} & 34.5 & 3.1 & 53.4 & 2.7 & 25.0 & 58.8 & 24.5 & 89.9 & 43.7 & 50.8 & 38.7 \\

 & centroids & \textcolor{MyGreen}{\cmark} & \textbf{74.9} & 51.8 & 81.8 & 81.3 & 88.2 & 88.6 & 98.5 & 96.6 & 67.3 & 83.8 & 81.3 \\
 & \cellcolor{LightGray}SiM (ours) & \cellcolor{LightGray}\textcolor{MyGreen}{\cmark} &  \cellcolor{LightGray}74.3 & \cellcolor{LightGray}\textbf{52.8} & \cellcolor{LightGray}\textbf{90.1} & \cellcolor{LightGray}\textbf{84.0} & \cellcolor{LightGray}\textbf{89.6} & \cellcolor{LightGray}\textbf{93.5} & \cellcolor{LightGray}\textbf{98.8} & \cellcolor{LightGray}\textbf{97.3} & \cellcolor{LightGray}\textbf{73.1} & \cellcolor{LightGray}\textbf{85.7} & \cellcolor{LightGray}\textbf{83.9} \\

\bottomrule

\end{tabular}}
\end{subtable}
\end{table*}

\begin{table*}[t]
\caption{Detailed results for the 10 datasets for two CLIP backbones. Top-1 accuracy averaged over 3 random seeds is reported. Embeddings $\boldsymbol{t}_k$ are obtained from imagenet classes, imagenet images (images) or a subset of 16452 words from Wordnet (Wordnet).}
\label{tab:other_prompts}
\centering
\begin{subtable}{0.85\linewidth}
\label{tab:other_prompts_vitb16}
\caption{Results for the CLIP ViT-B/16 backbone.}
\resizebox{\textwidth}{!}{%
\begin{tabular}{llcccccccccccc}
\toprule
 & Method & Vocabulary-free & SUN & Aircraft & EuroSAT & Cars & Food & Pets &  Flowers & Caltech & DTD & UCF & Average
\\ \midrule 
 & CLIP  & \textcolor{MyRed}{\xmark} & 62.6 & 24.7 & 47.5 & 65.3 & 86.1 & 89.1 & 71.4 & 92.9 & 43.6 & 66.7 &  65.0 \\

\midrule
\midrule
\multirow{3}{*}{\rotatebox{90}{\footnotesize 4-shot}}

&  SiM (images) & \textcolor{MyGreen}{\cmark} & 61.9 & 29.7 & 71.5 & 56.4 & 73.6 & 66.1 & 87.3 & 92.1 & 55.7 & 71.2 & 66.6 \\
&  SiM (Wordnet) & \textcolor{MyGreen}{\cmark} & 62.2 & 30.7 & 74.2 & \textbf{60.8} & \textbf{75.4} & \textbf{79.8} & \textbf{89.9} & 92.8 & 58.9 & 72.2 & 69.7 \\
 &\cellcolor{LightGray}SiM & \cellcolor{LightGray}\textcolor{MyGreen}{\cmark} & \cellcolor{LightGray}\textbf{62.7} & \cellcolor{LightGray}\textbf{33.2} & \cellcolor{LightGray}\textbf{75.8} & \cellcolor{LightGray}60.5 & \cellcolor{LightGray}\textbf{75.4} & \cellcolor{LightGray}79.2 & \cellcolor{LightGray}89.7 & \cellcolor{LightGray}\textbf{93.2} & \cellcolor{LightGray}\textbf{59.0} & \cellcolor{LightGray}\textbf{73.8} &  \cellcolor{LightGray}\textbf{70.2}\\
\midrule
\multirow{3}{*}{\rotatebox{90}{\footnotesize 8-shot}}

&  SiM (images) & \textcolor{MyGreen}{\cmark} & 67.1 & 37.5 & 70.2 & 65.4 & 79.2 & 77.8 & 92.4 & 94.7 & 61.9 & 75.6 & 73.1 \\
&  SiM (Wordnet) & \textcolor{MyGreen}{\cmark} & 66.9 & 37.4 & \textbf{81.0} & \textbf{69.1} & \textbf{80.4} & \textbf{84.8} & \textbf{93.6} & 94.4 & \textbf{66.0} & 76.3 & \textbf{75.0} \\
 &\cellcolor{LightGray}SiM & \cellcolor{LightGray}\textcolor{MyGreen}{\cmark} & \cellcolor{LightGray}\textbf{67.2} & \cellcolor{LightGray}\textbf{38.4} & \cellcolor{LightGray}80.1 & \cellcolor{LightGray}68.9 & \cellcolor{LightGray}80.1 & \cellcolor{LightGray}84.6 & \cellcolor{LightGray}92.8 & \cellcolor{LightGray}\textbf{94.8} & \cellcolor{LightGray}64.5 & \cellcolor{LightGray}\textbf{77.0}
 & \cellcolor{LightGray}74.9 \\
\midrule

\multirow{3}{*}{\rotatebox{90}{\footnotesize 16-shot}}

&  SiM (images) & \textcolor{MyGreen}{\cmark} & 70.0 & \textbf{44.2} & \textbf{85.7} & 73.4 & 82.2 & 83.9 & 94.8 & 95.3 & 69.6 & \textbf{79.6} & 77.9 \\
&  SiM (Wordnet) & \textcolor{MyGreen}{\cmark} & \textbf{70.3} & 44.0 & \textbf{85.7} & 74.7 & \textbf{82.8} & \textbf{88.1} & \textbf{95.5} & \textbf{95.7} & \textbf{70.8} & 79.1 & \textbf{78.7} \\
 &\cellcolor{LightGray}SiM & \cellcolor{LightGray}\textcolor{MyGreen}{\cmark} &  \cellcolor{LightGray}70.1 & \cellcolor{LightGray}43.7 & \cellcolor{LightGray}85.6 & \cellcolor{LightGray}\textbf{74.8} & \cellcolor{LightGray}\textbf{82.8} & \cellcolor{LightGray}\textbf{88.1} & \cellcolor{LightGray}\textbf{95.5} & \cellcolor{LightGray}95.3 & \cellcolor{LightGray}69.9 & \cellcolor{LightGray}79.0 &  \cellcolor{LightGray}78.5 \\
% \midrule
% \multirow{3}{*}{\rotatebox{90}{\footnotesize 32-shot}}

% &  SiM & \textcolor{MyGreen}{\cmark} & 71.6 & 49.2 & 89.3 & 78.0 & 84.0 & 90.4 & 96.4 & 95.7 & 72.9 & 81.5 & 80.9    \\
% &  SiM (images) & \textcolor{MyGreen}{\cmark} & & & & & & & & & & & \\
% &  SiM (Wordnet) & \textcolor{MyGreen}{\cmark} & 71.9 & 49.6 & 89.4 & 78.7 & 84.0 & 90.1 & 96.8 & 95.4 & 74.0 & 81.7 & 81.2 \\
\bottomrule
\end{tabular}}
\vspace{0.1cm}
\end{subtable}
\begin{subtable}{0.85\linewidth}
\label{tab:other_prompts_vitl14}
\caption{Results for the CLIP ViT-L/14 backbone.}
\resizebox{\textwidth}{!}{%
\begin{tabular}{llcccccccccccc}
\toprule
 & Method & Vocabulary-free & SUN & Aircraft & EuroSAT & Cars & Food & Pets &  Flowers & Caltech & DTD & UCF & Average
\\ \midrule 
 & CLIP  & \textcolor{MyRed}{\xmark} &  67.6 & 32.6 & 58.0 & 76.8 & 91.0 & 93.6 & 79.4 & 94.9 & 53.6 & 74.2 & 72.2 \\

\midrule
\midrule
\multirow{3}{*}{\rotatebox{90}{\footnotesize 4-shot}}

&  SiM (images) & \textcolor{MyGreen}{\cmark} & 67.0 & 39.5 & 79.6 & 70.7 & 84.5 & 80.7 & 95.5 & 95.0 & 59.1 & 80.3 & 75.2 \\
&  SiM (Wordnet) & \textcolor{MyGreen}{\cmark} & 66.8 & 39.1 & \textbf{82.6} & \textbf{73.2} & 85.5 & \textbf{87.8} & \textbf{96.3} & 95.6 & \textbf{62.5} & 79.8 & 76.9 \\
 &\cellcolor{LightGray}SiM & \cellcolor{LightGray}\textcolor{MyGreen}{\cmark} & \cellcolor{LightGray}\textbf{67.2} & \cellcolor{LightGray}\textbf{43.0} & \cellcolor{LightGray}80.9 & \cellcolor{LightGray}72.5 & \cellcolor{LightGray}\textbf{85.6} & \cellcolor{LightGray}87.3 & \cellcolor{LightGray}\textbf{96.3} & \cellcolor{LightGray}\textbf{95.8} & \cellcolor{LightGray}61.4 & \cellcolor{LightGray}\textbf{81.8} &  \cellcolor{LightGray}\textbf{77.2} \\
\midrule
\multirow{3}{*}{\rotatebox{90}{\footnotesize 8-shot}}

&  SiM (images) & \textcolor{MyGreen}{\cmark} & 71.7 & 47.3 & 84.5 & 76.9 & 87.5 & 87.4 & 97.0 & 97.3 & 66.5 & 83.0 & 79.9 \\
&  SiM (Wordnet) & \textcolor{MyGreen}{\cmark} & \textbf{71.8} & 45.7 & \textbf{87.1} & \textbf{79.8} & 88.3 & \textbf{91.7} & \textbf{98.3} & \textbf{96.9} & \textbf{70.0} & \textbf{83.8} & \textbf{81.4} \\
 & \cellcolor{LightGray}SiM & \cellcolor{LightGray}\textcolor{MyGreen}{\cmark} & \cellcolor{LightGray}71.6 & \cellcolor{LightGray}\textbf{47.6} & \cellcolor{LightGray}85.2 & \cellcolor{LightGray}\textbf{79.8} & \cellcolor{LightGray}\textbf{88.4} & \cellcolor{LightGray}91.6 & \cellcolor{LightGray}97.5 & \textbf{96.9} & \cellcolor{LightGray}68.3 & \cellcolor{LightGray}83.3 & \cellcolor{LightGray}81.0 \\
\midrule
\multirow{3}{*}{\rotatebox{90}{\footnotesize 16-shot}}

&  SiM (images) & \textcolor{MyGreen}{\cmark} & 74.0 & \textbf{53.0} & \textbf{90.3} & 82.9 & 89.1 & 91.7 & 98.5 & 96.8 & 71.3 & \textbf{86.2} & 83.4 \\
&  SiM (Wordnet) & \textcolor{MyGreen}{\cmark} & \textbf{74.7} & 52.9 & 90.0 & \textbf{84.1} & \textbf{89.7} & \textbf{93.8} & \textbf{98.9} & \textbf{97.5} & \textbf{74.2} & 85.9 & \textbf{84.2} \\
 &\cellcolor{LightGray}SiM & \cellcolor{LightGray}\textcolor{MyGreen}{\cmark} &  \cellcolor{LightGray}74.3 & \cellcolor{LightGray}52.8 & \cellcolor{LightGray}90.1 & \cellcolor{LightGray}84.0 & \cellcolor{LightGray}89.6 & \cellcolor{LightGray}93.5 & \cellcolor{LightGray}98.8 & \cellcolor{LightGray}97.3 & \cellcolor{LightGray}73.1 & \cellcolor{LightGray}85.7 &   \cellcolor{LightGray}83.9 \\
% \midrule
% \multirow{3}{*}{\rotatebox{90}{\footnotesize 32-shot}}

% &  SiM & \textcolor{MyGreen}{\cmark} &   &  & &  & & & & & & &   \\
% &  SiM (images) & \textcolor{MyGreen}{\cmark} & & & & & & & & & & &  \\
% &  SiM (Wordnet) & \textcolor{MyGreen}{\cmark} & 76.2 & 58.9 & 92.8 & 86.3 & 90.3 & 94.4 & 99.3 & 97.4 & 78.0 & 87.2 & 86.1 \\
%& $\Delta$ & & \textcolor{MyGreen}{+46.2} & \textcolor{MyGreen}{+40.8} & \textcolor{MyGreen}{+52.2} & \textcolor{MyGreen}{+69.5} & \textcolor{MyGreen}{+64.2} & \textcolor{MyGreen}{+16.3} & \textcolor{MyGreen}{+81.4} & \textcolor{MyGreen}{+17.8} & \textcolor{MyGreen}{+39.6} & \textcolor{MyGreen}{+35.7} & \\
\bottomrule

\end{tabular}}
\end{subtable}
\end{table*}

\section{Results}

\paragraph{Existing label mapping techniques are insufficient.} 
Table~\ref{tab:main_results} shows that state-of-the-art label mapping strategies consistently underperform compared to SiM across all 10 datasets. One-to-One mapping performs worse than BLM on average, highlighting the benefit of mapping multiple generic classes to each target category rather than relying on a rigid one-to-one correspondence. Surprisingly, the Centroids baseline—despite being a simple vision-only approach—largely outperforms BLM, raising questions about the viability of BLM in the few-shot learning setting for VLMs. When comparing BLM to SiM, we observe that Pets and Caltech101 exhibit the smallest performance gap, likely because their concepts are closely aligned with those in ImageNet. Conversely, Aircraft, Cars, and Flowers experience the largest performance drop, likely due to their fine-grained nature, which is either absent from ImageNet or only represented at a super-class level, making discrimination more challenging. 

% \begin{figure}
%     \centering
%     \includegraphics[width=\linewidth]{figures/visu.pdf}
%     \caption{Figure depicts importance weights after learning on 16-shot. Our approach allows to interpret final predictions based on generic classes/concepts.}
%     \label{fig:visu-results}
% \end{figure}

\begin{figure*}[t]
    \centering
    \begin{subfigure}[b]{0.9\textwidth}
        \centering
        \includegraphics[width=\textwidth]{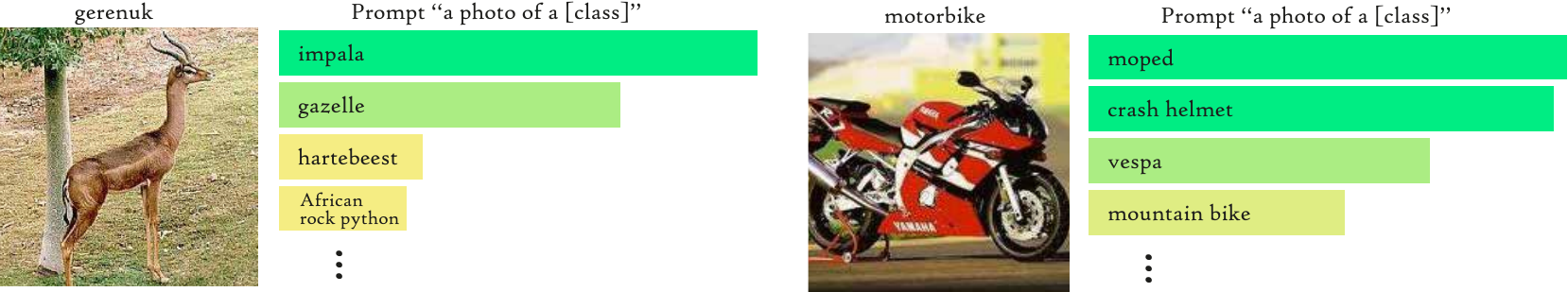}
        \caption{With ImageNet text prompts (``a photo of a \{$\text{class}$\}") on two classes of the Caltech101 dataset (gerenuk and motorbike).}
        \label{fig:visu_sub1}
    \end{subfigure}
    
    \vspace{0.5cm} % Space between subfigures

    \begin{subfigure}[b]{0.9\textwidth}
        \centering
        \includegraphics[width=\textwidth]{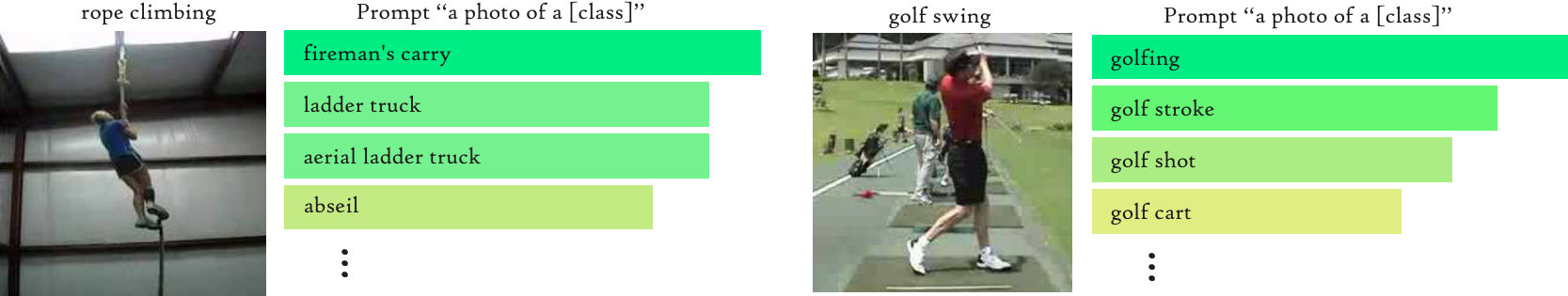}
        \caption{With Wordnet text prompts (``a photo of a \{$\text{class}$\}") on two classes of the UCF101 dataset (rope climbing and golf swing).}
        \label{fig:visu_sub2}
    \end{subfigure}

    \vspace{0.5cm} % Space between subfigures

    \begin{subfigure}[b]{0.9\textwidth}
        \centering
        \includegraphics[width=\textwidth]{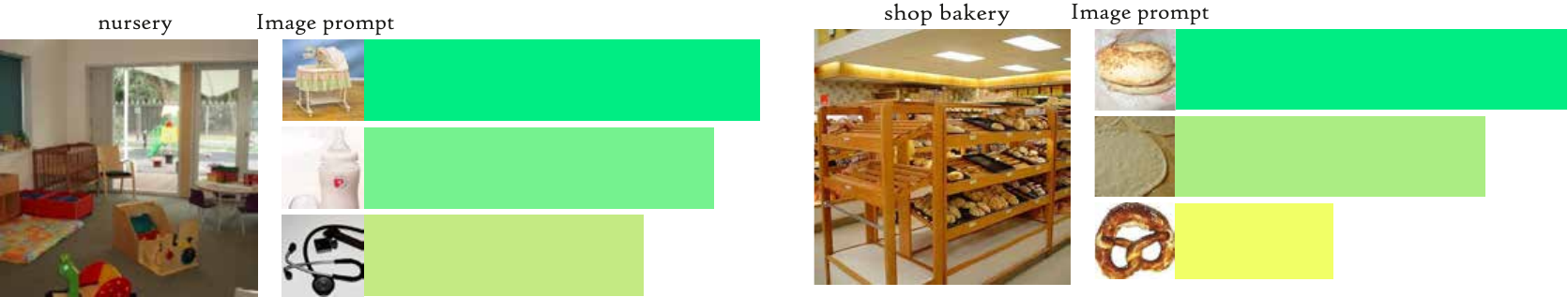}
        \caption{With ImageNet image prompts (one image per class) on two classes of the SUN397 dataset (nursery and shop bakery).}
        \label{fig:visu_sub3}
    \end{subfigure}
    
    \caption{The linear mapping is learned in the 16-shot setting with the ViT-L/14 backbone. We order the generic prompts $k$ according to the values of the weights $w_{k,c}$ for the given target class $c$, and retain the four highest. The heights of the bars are proportional to these values, while their color range from green (highest) to yellow (lowest).}
    \label{fig:visu}
\end{figure*}
\paragraph{SiM is competitive with zero-shot classification.} Table~\ref{tab:main_results} presents zero-shot results, which are only achievable when class names are available. Our approach outperforms zero-shot CLIP on 6 out of 10 datasets with 4 shots per class and on 8 out of 10 datasets with 16 shots per class. However, performance varies significantly across datasets. On datasets such as Flowers, DTD, and EuroSAT, SiM surpasses zero-shot CLIP with only 4 shots per class, whereas for Food101 and Pets, SiM remains below zero-shot performance even in the 16-shot setting. These results suggest that incorporating even partial or noisy prior knowledge—such as an initial zero-shot prediction from a captioning model~\cite{conti2024vocabulary}—could further enhance SiM’s performance, particularly in cases where vocabulary-free methods struggle to reach zero-shot accuracy.

%our approach offers interpretability by potentially 

%We can also see in the Ablation Study (Section \ref{}) some visualization which can help get in describing the unknown classes. This interpretability is a great advantage of our approach and can be seen as a first towards vocabulary-free classification, as discussed in [ref vocab-free classification].
\begin{table}[t]
\caption{Additional results on ImageNet. Top-1 accuracy averaged over 3 random seeds is reported. For SiM, embeddings $\boldsymbol{t}_k$ are obtained from a subset of 16452 words from Wordnet.}
\label{tab:imagenet_Wordnet}
\centering

\begin{subtable}{0.95\linewidth}
 \renewcommand{\arraystretch}{0.8}
\label{tab:imagenet_vitb16}
\caption{Results for the CLIP ViT-B/16 backbone. For reference, the zero-shot (not vocabulary-free) performance is $66.7$.}
\resizebox{\textwidth}{!}{%
\begin{tabular}{l|cccc}
\toprule
 \footnotesize{Method} & \footnotesize{4-shot} & \footnotesize{8-shot} & \footnotesize{16-shot} & \footnotesize{32-shot} %&  \footnotesize{64 shots} 
 \\ \midrule   
 %\footnotesize{One-to-One} & & & & \\\footnotesize{BLM} & & & & \\
 \footnotesize{Centroids} & \footnotesize{42.1} & \footnotesize{47.8} & \footnotesize{51.0} & \footnotesize{53.9} %& \footnotesize{55.8}
 \\
 \rowcolor{LightGray}\footnotesize{SiM (Wordnet)} & \textbf{\footnotesize{52.5}} & \textbf{\footnotesize{57.8}} & \textbf{\footnotesize{61.0}} &
 \textbf{\footnotesize{62.9}} %& 
 \\
  %Vocabulary-free   & \textcolor{MyGreen}{\cmark} & \textcolor{MyGreen}{\cmark} \\
\bottomrule

\end{tabular}}

\end{subtable}

\vspace{10pt}

\begin{subtable}{0.95\linewidth}
 \renewcommand{\arraystretch}{0.8}
\label{tab:imagenet_vitl14}
\caption{Results for the CLIP ViT-L/14 backbone. For reference, the zero-shot (not vocabulary-free) performance is $75.9$.}
\resizebox{\textwidth}{!}{%
\begin{tabular}{l|cccc}
\toprule
 \footnotesize{Method} & \footnotesize{4-shot} & \footnotesize{8-shot} & \footnotesize{16-shot} & \footnotesize{32-shot} %&  \footnotesize{64 shots} 
 \\ \midrule   
 %\footnotesize{One-to-One} & & & & \\\footnotesize{BLM} & & & & \\
 \footnotesize{Centroids} & \footnotesize{50.9} & \footnotesize{57.2} & \footnotesize{60.7} &
 \footnotesize{63.2} 
 %& \footnotesize{65.6}
 \\
 \rowcolor{LightGray}\footnotesize{SiM (Wordnet)} & \textbf{\footnotesize{63.9}} & \textbf{\footnotesize{68.7}} & \textbf{\footnotesize{71.2}} &
 \textbf{\footnotesize{72.8}} %& \footnotesize{} 
 \\
  %Vocabulary-free   & \textcolor{MyGreen}{\cmark} & \textcolor{MyGreen}{\cmark} \\
\bottomrule

\end{tabular}}
\end{subtable}
\end{table}

\paragraph{Lack of vocabulary degrades performances.} Despite promising results, Table~\ref{tab:main_results} shows that a performance gap remains between SiM and traditional few-shot learning methods, which benefit from explicit class names. However, this gap significantly decreases when using a more powerful backbone such as ViT-L/14, particularly for datasets like Flowers, Caltech101, and EuroSAT. We could also hypothesize that combining recent observations on current few-shot learning for VLMs with our approach (e.g., combination with prompt tuning, adapter or low-rank adaptation methods) could help bridge the remaining gap.

\paragraph{Images can be used as prompts but lag behind textual prompts.}
Table \ref{tab:other_prompts} show results obtained with different prompts. When the embeddings $\boldsymbol{t}_{k}$ are obtained from a single image from the classes of ImageNet, performances tend to be slightly lower than when using ImageNet class names. This gap narrows when the number of shots increases, or when using a more powerful backbone. In the 16-shot setting with ViT-L/14, image-based and text-based prompts yield comparable performance on most datasets.

\paragraph{Textual prompts are not restricted to ImageNet classes.}
Table \ref{tab:other_prompts} show results obtained with different prompts, namely a subset of $K=16452$ words from Wordnet to generate the embeddings. Our approach scales gracefully as learning the mapping with 16-shot ImageNet ($N = 16000$) takes 0.8s on a Tesla A100. The performances slightly degrade compared to the ImageNet classes in the 4-shot setting for both backbones. This could be due to an increased sensitiveness to noise because of the much larger $K$, where the learned mapping generalizes not as effectively at test time. In the 8 and 16-shot settings, performances are on par on average. In this setting, we can also provide results for ImageNet, presented in Table \ref{tab:imagenet_Wordnet}. Similar to some other datasets, e.g. Food, the performance of SiM remains below zero-shot even with 16 and 32 shots. The gap is narrower with the more powerful ViT-L/14 backbone.

\paragraph{Learned weights can provide semantic understanding.} Beyond classification performance, we can investigate the relative importance of weights $w_{k,c}$ for a given target class $c$ (see Eq.~\eqref{eq:target_similarities}) which potentially provide a high level semantic understanding. Figure~\ref{fig:visu} illustrates this interpretability across the three types of prompts studied in this work. Figure~\ref{fig:visu_sub1} presents results for ImageNet class names prompts on Caltech101 (animals and objects), where our approach meaningfully associates gerenuk with impala and gazelle, other antelope species. Figure~\ref{fig:visu_sub2} shows results for the UCF101 dataset (action recognition) using Wordnet-based prompts, where semantically related words to rope climbing such as abseil appear as relevant matches. Figure~\ref{fig:visu_sub3} presents results for the SUN397 dataset (scene recognition) using image-based prompts, demonstrating that a shop bakery is linked to baked goods such as bagels, dough, and pretzels. 
\begin{figure}[t]

    \centering
    \begin{subfigure}[b]{0.9\linewidth}
        \centering
        \includegraphics[width=\textwidth]{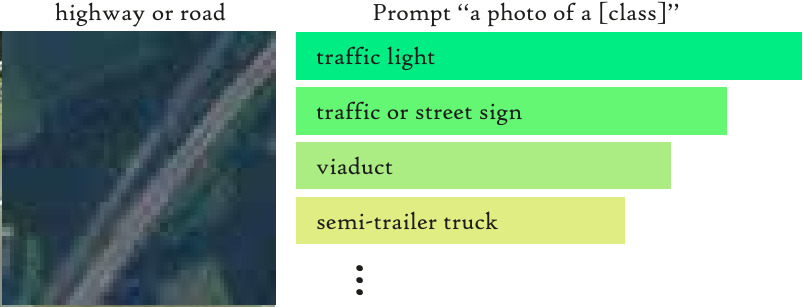}
        \caption{An ambiguous example from the EuroSAT dataset with target class ''Highway or road", linked to semantically related concepts which cannot be seen on the visual shots.}
        \label{fig:visu_highway}
    \end{subfigure}
    
    \vspace{0.3cm} % Space between subfigures

    \begin{subfigure}[b]{0.9\linewidth}
        \centering
        \includegraphics[width=\textwidth]{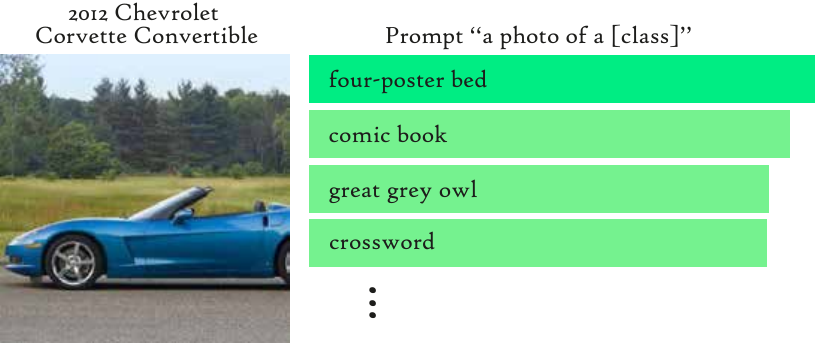}
        \caption{An ambiguous example from the StanfordCars dataset with target class ''2012 Chevrolet Corvette Convertible", for which no meaningful generic prompt can be matched.}
        \label{fig:visu_chevrolet}
    \end{subfigure}
\caption{The linear mapping is learned in the 16-shot setting with the ViT-L/14 backbone. We order the generic prompts $k$ according to the values of the weights $w_{k,c}$ for the given target class $c$, and retain the four highest. The heights of the bars are proportional to these values, while their color range from green (highest) to yellow (lowest).}
\label{fig:visu_bad}
\end{figure}

\paragraph{The linear mapping is not always interpretable.} The information contained in the learned mapping is not straightforwardly interpretable for every target classes. For instance, Figure \ref{fig:visu_highway} shows an example where the most important weights are associated with generic prompts semantically related to the target class, but which do not appear in the images defining it. %This highlights how the information contained in the learned mapping may be ambiguous. 
An other interpretability failure case appears to be linked with fine-grained datasets, as shown in Figure \ref{fig:visu_chevrolet}, where the target class is matched with completely unrelated concepts. Interestingly, this lack of interpretability does not seem to be detrimental to accuracy, as SiM achieves a 31\% reduction of the error rate in the 16-shot setting with the ViT-L/14 backbone (i.e., the setting with which Figure \ref{fig:visu_chevrolet} was obtained) compared to zero-shot classification with class names. Therefore, the choice of generic prompts may induce a trade-off between interpretability and separability of target classes.

\section{Discussion and future works}
As only few works have been targeting the problem of vocabulary-free image classification so far, the avenues for future research seem numerous in vocabulary-free few-shot learning for VLMs. The results presented in Table~\ref{tab:main_results} suggest there may be opportunities for adapting recent advances in few-shot adaptation for VLM, such as prompt tuning, to this novel setting. Furthermore, our baseline could be tweaked by proposing a different regularization, to promote higher classification accuracy or better interpretability. 

Beyond refining the learning algorithm, enhancing the choice of generic prompts could also play a crucial role in improving performance. As seen in Table~\ref{tab:other_prompts}, the set of generic prompts may come from very diverse sources, suggesting potential for optimizing the choice of the embeddings $\boldsymbol{t}_k$ used to compute the similarities. This in turn could be helpful to mitigate issues illustrated in Figure \ref{fig:visu_chevrolet}.

Finally, associating meaningful names with groups of shots remains an open challenge. For example, by investigating techniques for automatically labeling discovered classes while evaluating performance with semantic metrics, such as semantic intersection over union (IoU), as suggested in~\cite{conti2024vocabulary}.

\section{Conclusion}
In this work, we introduced {\em vocabulary-free few-shot learning}, a new framework for image classification using VLMs where class names are not available. We highlighted how current few-shot adaptation methods are ill-equipped to handle this practical scenario. To address this limitation, we proposed a simple yet effective baseline that leverages similarity scores between few-shot images and a set of generic, arbitrary prompts, which can be sourced from texts or images, predefined without any knowledge of the target classes. Interestingly, our least-squares baseline, SiM, achieves performance close to that of more complex few-shot adaptation techniques that rely on explicit class names. Additionally, SiM does not require direct access to the embeddings of either the generic prompts or the images—only the similarity scores. Finally, we demonstrated that analyzing the relative weights of the learned linear mapping could potentially provide high-level semantic insights into the target classes. Overall, we hope this work serves as a stepping stone for {\em vocabulary-free few-shot learning}, an important yet overlooked problem in VLM adaptation.

%which may arise when classes are not easily described or when groups of images are obtained from an unsupervised clustering algorithm
%\paragraph{Future works.} 

%\begin{itemize}
   % \item other text promps such as wordnet or use external image-text pairs datasets.
   % \item other type of prompts such as images
   % \item more complex mapping (combining recent advances in VLM few-shot learning such as prompt tuning)
   % \item enhance interpretability of final prediction, e.g., by stronger regularization
   % \item adapt existing few-shot methods to vocabulary-free setting
   % \item make link with base to novel setting
%\end{itemize}

%The linear mapping is learned in the 16-shot setting with the ViT-L/14 backbone. We order the generic prompts $k$ according to the values of the weights $w_{k,c}$ for the given target class $c$, and retain the four highest. The heights of the bars are proportional to these values, while their color range from green (highest) to yellow (lowest).
\section{Acknowledgments}
M.~Zanella is funded by the Walloon region under grant No.~2010235 (ARIAC by DIGITALWALLONIA4.AI). C.~Fuchs is funded by the MedReSyst
project, supported by FEDER and the Walloon Region. Part of the computational resources have been provided by the Consortium des Équipements de Calcul Intensif (CÉCI), funded by the Fonds de la Recherche Scientifique de Belgique (F.R.S.-FNRS) under Grant No. 2.5020.11 and by the Walloon Region.

{
    \small
    \bibliographystyle{ieeenat_fullname}
    \bibliography{main}
}

%\input{sections/appendix}
\end{document}